\pdfoutput=1

\documentclass[11pt]{article}

\usepackage{authblk} 
\usepackage[final]{acl}

\usepackage{times}
\usepackage{latexsym}

\usepackage[T1]{fontenc}

\usepackage[utf8]{inputenc}

\usepackage{microtype}

\usepackage{inconsolata}

\usepackage{graphicx}

\usepackage{multirow}
\usepackage{array}
\usepackage{makecell}
\usepackage{amsmath}
\usepackage{hyperref}
\usepackage[inline]{enumitem}
\usepackage{soul}
\usepackage{xcolor}
\usepackage{tabularray}
\usepackage{floatrow}
\usepackage{ragged2e}
\DeclareFloatFont{tiny}{\tiny}
\usepackage{color}
\usepackage[normalem]{ulem}
\usepackage{rotating}

\title{Provenance: A Light-weight Fact-checker for \\ Retrieval Augmented LLM Generation Output}

\author{\textbf{Hithesh Sankararaman}\thanks{hithesh.sankararaman@uniphore.com}}
\author{\textbf{Mohammed Nasheed Yasin}\thanks{mohammed.yasin@uniphore.com}}
\author{\\ \textbf{Tanner Sorensen}}
\author{\textbf{Alessandro Di Bari}}
\author{\textbf{Andreas Stolcke}}

\affil{Uniphore Software Systems Pvt. Ltd, U.S.A}

\begin{document}
\maketitle
\begin{abstract}
We present a light-weight approach for detecting nonfactual outputs from retrieval-augmented generation (RAG). Given a context and putative output, we compute a factuality score that can be thresholded to yield a binary decision to check the results of LLM-based  question-answering, summarization, or other systems. Unlike factuality checkers that themselves rely on LLMs, we use compact, open-source natural language inference (NLI) models that yield a freely accessible solution with low latency and low cost at run-time, and no need for LLM fine-tuning. The approach also enables downstream mitigation and correction of hallucinations, by tracing them back to specific context chunks. Our experiments show high area under the ROC curve (AUC) across a wide range of relevant open source datasets, indicating the effectiveness of our method for fact-checking RAG output.
\end{abstract}

\section{Introduction}

With natural language understanding applications increasingly relying on large language models (LLMs) to answer questions, summarize texts, and perform other tasks, detecting nonfactual claims in the generated text has become critical from an ethical and compliance standpoint. LLMs, while powerful, are prone to generate nonfactual or ``hallucinated'' information that can lead to misinformation and introduce errors in business processes. To address this problem, we present \textit{Provenance}, a fact-checking method for output generated by LLMs, with respect to a given context that provides the factual basis for the output.\\

\textit{Provenance} leverages compact cross-encoder models that offer substantial advantages over conventional LLM-based methods. These advantages include accessibility, low latency/high throughput, and interpretable judgments.

\textit{Provenance} is evaluated on diverse open-source datasets, including the TRUE dataset \cite{honovich2022true}, MSMarco \cite{DBLP:journals/corr/NguyenRSGTMD16}, TruthfulQA \cite{lintruthfulqa}, HotpotQA \cite{yang2018hotpotqadatasetdiverseexplainable}, HaluEval \cite{li2023halueval} and HaluBench \cite{ravi2024lynxopensourcehallucination}. These datasets encompass a variety of question-answering contexts, providing a robust testbed for our methods. We assess performance using standard detection metrics to demonstrate our method's efficacy as a factuality checker for LLM-generated content.

Our findings show that \textit{Provenance} achieves competitive hallucination detection performance (as measured by AUC) across different datasets, thus contributing to improved trustworthiness and utility of LLMs in real-world applications.

\section{Related Work}

In prior work, three main approaches to factuality evaluation have been used:
\begin{enumerate*}
    \item LLM ablation,
    \item LLM introspection, and
    \item NLI methods.
\end{enumerate*}

\textit{LLM ablation} refers to approaches such as SelfCheckGPT \cite{manakul-etal-2023-selfcheckgpt} and \citet{agrawal-etal-2024-language} that measure the consistency of multiple candidate generations for a given prompt. Methods such as \citet{varshney2023stitchtimesavesnine} that gauge factuality based on the language model's output distributions also fall in this category.

\textit{LLM introspection} refers to techniques that use the reasoning ability of modern language models to evaluate their own or another model's output. Work by \citet{kadavath2022languagemodelsmostlyknow}, \citet{es-etal-2024-ragas} and \citet{muller-etal-2023-evaluating} are examples of this.

\textit{Natural language inference} (NLI) methods exploit special-purpose cross-encoder models that indicate whether a claim is supported by a premise. This approach usually involves breaking down the context into a list of premises (\textit{context items}), and the generation into a list of claims. \citet{laban-etal-2022-summac} is a representative method that chunks the generation and context at the sentence level and computes pair-wise entailment judgments, which are then aggregated. However, this approach has some shortcomings: 
\begin{enumerate*}
    \item the original prompt/query is ignored when evaluating entailment, and
    \item context and generation chunking is overly simplistic.
\end{enumerate*}
Our method falls into the NLI-based category, but addresses these shortcomings.

\begin{figure}[tb]
  \includegraphics[width=\linewidth]{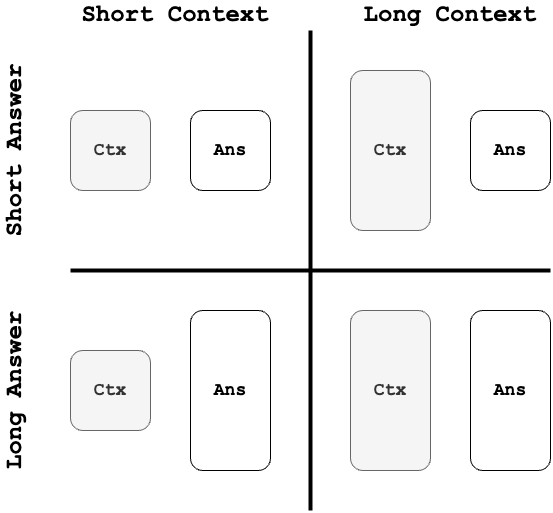}
  \caption {Typical Context vs.\ Answer length scenarios in which fact-checking is performed.}
   \label{fig:scope}
\end{figure}

Broadly speaking, there are four scenarios (Fig.~\ref{fig:scope}) in which a fact-checker may operate:
\begin{enumerate*}
    \item short context/short answer,
    \item short context/short answer,
    \item short context/long answer, and
    \item long context/long answer.
\end{enumerate*}
When the answer or context are long, we need a mechanism to break them into smaller units. We narrow our focus based on the following observations and practical considerations:
\begin{enumerate}
    \item Reliable semantic chunking is an as yet evolving field in NLP \cite{yang2020we,zhai2017neural,johnson2005high}.
    \item When it comes to chunking long contexts we can reuse the chunks that the RAG \textit{retriever} returned. Retrievers need to chunk text due to input sequence length limitations in their embedder.
    \item Lack of open-source datasets for long-answer benchmarking.
\end{enumerate}
While we have a straightforward way to break down the contexts, it is still hard to chunk generated answers meaningfully. The chunking of information is an area for further research, since context and answers come in many forms, such as text, conversations, and tables. We limit the scope of this paper to text source for scenarios in the first row of Figure \ref{fig:scope}, namely, short context/short answer and long context/short answer. We also need to ensure that the chunk length chosen is viable for all the models in the system.

\section{System Description} \label{System Description}

\begin{figure*}[t]
  \includegraphics[width=\textwidth]{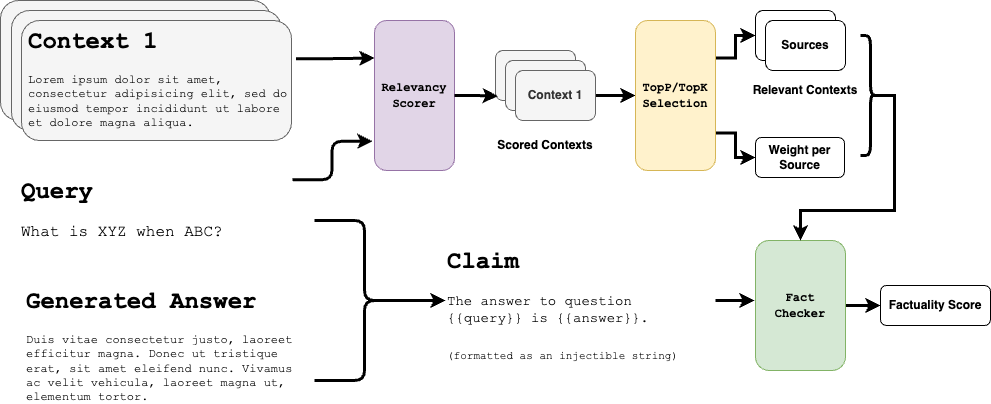}
  \caption {Provenance system architecture.}
   \label{fig:system-diagram}
\end{figure*}

Contemporary fact-checking systems employ approaches based on LLMs as a judge \cite{zhu2023judgelm} to validate the generations of other LLMs. By virtue of being auto-regressive, the judge-LLMs themselves are prone to hallucinate. By contrast, Provenance (Figure \ref{fig:system-diagram}) uses two cross-encoder based models that do not suffer from this tendency. As input, Provenance expects
\begin{enumerate}
    \item a list of context items used by the generating LLM in the upstream RAG,
    \item the user's original question or query, and
    \item the generated text to fact-check. 
\end{enumerate}

The first cross-encoder model determines which of the context items are relevant to the given query and generates a score. This score is then used to select context items to build a smaller and more focused context, which we refer to as the \textit{sources}. The selection process also produces a \textit{weight} associated with each source.
In parallel. we construct the \textit{claim} by inserting the query and generation into a \textit{claim prompt}. The \textit{claim} and \textit{sources} are then passed to the second cross-encoder model for validation, generating a \textit{factuality score} for each \textit{claim/source} pair.
These scores are then aggregated using the source \textit{weights} generated earlier to produce a single score for the LLM's output. This score can be thresholded to produce a binary factuality decision, with the threshold being tuned for a target dataset and task.  Here we used threshold-invariant evaluation methods, such as receiver operator characteristics (ROC) and area under the curve (AUC).

\subsection{Relevancy Scorer} \label{reranker-method}

To assess the relevance of context items to the query, we use a cross-encoder model to generate relevance scores for each context item. This process is similar to the re-ranking of search results w.r.t.\ queries in a RAG system, except that we do not perform the \textit{top\_k} sampling step. We leave this to a downstream component.\\
Given a query $Q$ and a context item $D$, the relevance score $S$ is calculated as
\begin{equation}
    S = \text{Cross-Encoder}(Q, D)
\end{equation}
Here, $S$ is a real number in $(-\infty, \infty)$, but empirically scors range within $(-10, 10)$.

The cross-encoder used is a RoBERTa-based model\footnote{Available on huggingface as \href{https://huggingface.co/mixedbread-ai/mxbai-rerank-base-v1}{mixedbread-ai/mxbai-rerank-base-v1}} trained by \href{https://www.mixedbread.ai/}{Mixedbread}.

\subsection{Context Item Selection}

To select the \textit{sources} among the scored contexts, we employ one of two strategies. \emph{TopK} is similar to the one used in the RAG retrieval and reranking steps.
\emph{TopP} is adapted from nucleus sampling \cite{holtzman2019curious}, a commonly used method to sample from an LLM's output distribution.
For both strategies, the relevance scores of all context items are normalized to be interpretable as probabilities, i.e., to have range $(0,1)$ and sum to one.

The TopK selector simply retains the \textit{top\_k} contexts with highest relevance scores.
The TopP selector retains a minimal set of contexts in order of decreasing relevance scores, such that their cumulative probability is at least \textit{top\_p}, where \textit{top\_k} and \textit{top\_p} respectively are hyperparameters.

Following the selection of the \textit{sources}, we re-normalize their relevance scores again, which then serve as the \textit{weights} to be placed on each source later in fact-checking.

We have not carried out a systematic optimization of \textit{top\_k} and \textit{top\_p} values for this paper.
For \textit{top\_p} we chose 0.9, which selected an average of 3 to 4 sources on our datasets.
For  \textit{top\_k} we chose 5, which is half the maximum of possible sources defined in the datasets used here (see Section \ref{data_prep}).

Anecdotally, on real-world production datasets, we found that better results are achieved by choosing a single \textit{top\_p} value rather than setting \textit{top\_k}.

\subsection{Fact Checker} \label{nli-model}

Provenance uses cross-encoder NLI models to evaluate the factual consistency of the LLM's output, given a \textit{source} and the user's query. The model we use is a specialized hallucination detection model\footnote{Available on Huggingface as \href{https://huggingface.co/vectara/hallucination_evaluation_model}{vectara/hallucination\_evaluation\_model}} trained by \href{https://vectara.com/}{Vectara}.

The steps to compute factuality scores are
\begin{enumerate}
    \item {\em Input preparation:} we insert the query and answer into a prompt that claims ``The answer to question <QUERY> is <ANSWER>.'' This prepared claim prompt is then paired off with each \textit{source}.
    \item {\em Scoring:} The cross-encoder is used to compute a score indicating how well the answer is supported by the context in the light of the query. Here, the scoring function is $FScore = \text{nli-model}(S, C)$,
    where $S$ is one of the sources and $C$ is the prepared claim prompt.
    \item {\em Aggregation} of the scores and weights for all the {\em sources} using one of the following functions: 
    \begin{enumerate*}
        \item min,
        \item max, or
        \item weighted average.
    \end{enumerate*}
\end{enumerate}
The final factuality score can be normalized to indicate the probability of the claim being supported by the \textit{sources}.

\begin{table*}[t!]
\centering
\caption{Comparison of AUC scores and model sizes from the TRUE paper with our Provenance framework; we report AUC scores*100 for better readability, as in the TRUE paper \cite{honovich2022true}. Results from FEVER, PAWS, and VITAMIN C (reported above, but crossed-out) are not comparable to the TRUE results since our NLI model has seen samples from these datasets. The highest score for our method is in bold with an asterisk, while the highest score from the TRUE paper methods is in bold. The size of the Provenance model is $\approx$ 300M parameters.}
\label{table:TRUE-accuracy-table}
\begin{tblr}{
  column{3} = {c},
  column{4} = {c},
  column{5} = {c},
  column{6} = {c},
  cell{3}{1} = {r=3}{},
  cell{6}{1} = {r=5}{},
  cell{11}{1} = {r=2}{},
  vlines,
  hline{1-3,6,11,13-14} = {-}{},
  hline{4-5,7-10,12} = {2-6}{},
}
\textbf{Data Type}                          & \textbf{Dataset}                              & \textbf{Sample Count} & {\textbf{AUC}\\ (Provenance)}  & {\textbf{AUC} \\ \textbf{(TRUE paper)}} & {\textbf{Model size}\\ \textbf{(TRUE paper)}} \\
{\textbf{Paraphrase} \\ \textbf{Detection}}               & \textbf{PAWS}                                 & 8000                  & \sout{\textbf{94*}} & 89.7$^{Q^{2}}$               & 11B              \\
{\textbf{Dialogue} \\ \textbf{Generation}}                 & \textbf{BEGIN}                                & 836                   & 80                   & \textbf{87.9$^{BERT\_SCORE}$} & 750M                               \\
                                             & \textbf{DialFact}                             & 8689                  & \textbf{92*}         & 86.1$^{Q^{2}}$              & 11B    \\
                                             & \textbf{Q2}                                   & 1088                  & \textbf{86*}         & 80.9$^{Q^{2}}$               & 11B   \\
{\textbf{Abstractive} \\ \textbf{Summarization}}          & \textbf{FRANK}                                & 671                   & 89                   & \textbf{89.4$^{ANLI}$}       & 11.5B                              \\
                                             & \textbf{MNBM}                                 & 2500                  & \textbf{79*}         & 77.9$^{ANLI}$               & 11.5B    \\
                                             & \textbf{QAGS\_CNNDM}                          & 235                   & 76.3                 & \textbf{83.5$^{Q^{2}}$}      & 11B                                \\
                                             & \textbf{QAGS\_XSUM}                           & 239                   & 80.4                 & \textbf{83.8$^{ANLI}$}       & 11.5B                              \\
                                             & \textbf{Summ\_Eval}                           & 1600                  & 70.1                 & \textbf{81.7$^{SC\_ZS}$}     & 58.7M~                             \\
\textbf{Fact Verification}                  & \textbf{VITAMIN C}                            & 63054                 & \sout{\textbf{95.8*}} & 88.3$^{ANLI}$               & 11.5B               \\
                                             & \textbf{FEVER}                                & 18209                 & \sout{92}            & 93.2$^{ANLI}$               & 11.5B                              \\
\end{tblr}
\end{table*}

\section{Data} \label{data}

We utilize several open-source datasets to evaluate the effectiveness of our approach in detecting nonfactual texts generated by LLMs. These datasets provide a diverse range of question-answering contexts and candidate answers, ensuring a comprehensive assessment. Table \ref{table:DD} provides an overview of datasets showing the counts of Hallucination and Entailment (=Factual) labels. 
As shown, most data sources have a roughly balanced label distribution, though some (like the HaluEval GENERAL subset) are skewed toward one class.

\subsection{TRUE}

The TRUE dataset \cite{honovich2022true} is comprised of eleven different subsets, each with questions, answers, and contexts. It is designed to test the factual accuracy of LLM outputs across various domains and question types.

\subsection{MSMarco}

MSMarco (Microsoft MAchine Reading COmprehension) \cite{DBLP:journals/corr/NguyenRSGTMD16} is a large-scale dataset created for machine reading comprehension tasks. The dataset is particularly useful for evaluating our method in the context of web-based information retrieval and answering user queries accurately.

\subsection{Truthful QA}

TruthfulQA \cite{lintruthfulqa} is a dataset specifically designed to test the truthfulness of LLM-generated responses. This dataset is crucial for assessing our approach's capability to handle tricky or potentially deceptive questions.

\subsection{HotpotQA}

HotpotQA \cite{yang2018hotpotqadatasetdiverseexplainable} is a multi-hop question-answering dataset that requires the model to retrieve and reason over multiple pieces of evidence to generate a correct answer. The dataset includes questions, supporting facts, and distractor contexts, making it a complex and rigorous test for our method. The multi-hop nature of HotpotQA ensures that our approach can handle intricate reasoning and context synthesis tasks effectively.

\subsection{HaluEval}

Hallucination Evaluation Benchmark for Large Language Models (HaluEval) \cite{li2023halueval} is a large collection of generated and human-annotated hallucinated samples for evaluating the performance of LLMs in recognizing hallucination. 

\subsection{HaluBench}

HaluBench \cite{ravi2024lynxopensourcehallucination} is a hallucination evaluation benchmark of 15k samples that consists of context-question-answer triplets annotated for whether the examples contain hallucinations. Compared to prior datasets, HaluBench is the first open-source benchmark containing hallucination tasks sourced from real-world domains that include finance and medicine.

\begin{table*}[t!]
\centering
\setlength{\tabcolsep}{5pt} 
\renewcommand{\arraystretch}{1.5} 
\begin{tblr}{|>{\centering\arraybackslash}p{2.5cm}|>{\centering\arraybackslash}p{3cm}|>{\centering\arraybackslash}p{2.5cm}|>{\centering\arraybackslash}p{2cm}|>{\centering\arraybackslash}p{2cm}|}
\hline
\textbf{Dataset Name} & \textbf{Sub Dataset Name} & \textbf{Label 0 (Hallucination)} & \textbf{Label 1 (Entailment)} & \textbf{Total Samples} \\ \hline
\SetCell[r=11]{c}{\textbf{TRUE}} & VITC & 31570 & 31484 & 63054 \\ \cline{2-5}
 & BEGIN & 554 & 282 & 836 \\ \cline{2-5}
 & DIALFACT & 5348 & 3341 & 8689 \\ \cline{2-5}
 & FEVER & 11816 & 6393 & 18209 \\ \cline{2-5}
 & FRANK & 448 & 223 & 671 \\ \cline{2-5}
 & MNBM & 2245 & 255 & 2500 \\ \cline{2-5}
 & PAWS & 4461 & 3539 & 8000 \\ \cline{2-5}
 & Q2 & 460 & 628 & 1088 \\ \cline{2-5}
 & QAGS\_CNNDM & 122 & 113 & 235 \\ \cline{2-5}
 & QAGS\_XSUM & 123 & 116 & 239 \\ \cline{2-5}
 & SUMMEVAL & 294 & 1306 & 1600 \\ \hline
\textbf{MS MARCO} &  & 252 & 252 & 504 \\ \hline
\textbf{HOTPOTQA} &  & 10000 & 100447 & 110447 \\ \hline
\textbf{HALUBENCH} &  & 7170 & 7730 & 14900 \\ \hline
\textbf{TRUTHFUL\_QA} &  & 1716 & 1260 & 2976 \\ \hline
\SetCell[r=4]{c}{\textbf{HALUEVAL}} & DIALOGUE & 10000 & 10000 & 20000 \\ \cline{2-5}
 & QA & 10000 & 10000 & 20000 \\ \cline{2-5}
 & SUMMARIZATION & 10000 & 10000 & 20000 \\ \cline{2-5}
 & GENERAL & 815 & 3692 & 4507 \\ \hline
\textbf{TOTAL} &  & \textbf{107394} & \textbf{191061} & \textbf{298455} \\ \hline
\end{tblr}
\caption{Overview of Datasets and Sub-Datasets Categorized by Hallucination and Entailment Labels, including Total Sample Counts. (Entailment corresponds to Factual for our purposes.)}
\label{table:DD}
\end{table*}

\section{Data Preparation}
    \label{data_prep}

 The MSMarco and HotpotQA datasets each contain 10 \textit{sources} per question, with one relevant \textit{source} per question in MSMarco and multiple relevant\textit{sources} per question in HotpotQA. Other datasets have a single \textit{source} paragraph given for each question. All \textit{sources} were split into individual sentences, and all datasets were converted into triplets with the query and answer as strings, and the \textit{sources} as a list of strings.Our framework processes these triplets and returns a score, which, combined with a set threshold, classifies the generated answer as hallucinated or factual. To calculate AUC, we ensured representation of the two classes by generating hallucinated answers for datasets lacking them.

For the MSMarco dataset \cite{DBLP:journals/corr/NguyenRSGTMD16}, we randomly selected 252 out of 100,000 datapoints and generated hallucinated answers using the GPT-3.5-turbo model, which were verified manually.

For HotpotQA \cite{yang2018hotpotqadatasetdiverseexplainable}, we appended the QA data from HaluEval \cite{li2023halueval}, which includes 10K hallucinated samples based on HotpotQA.

\section{Experiments}

\subsection{Preliminary Experiments}
Before developing our final Provenance framework, we also experimented with a BERT-based Relevancy Scorer using TopP selection 
and a DeBERTa-based NLI model for computing factuality scores.
These preliminary experiments showed the importance of (1) sorting of selected sources into their original temporal order and (2) cosine scoring (length normalization) of similarity scores; detailed results can be found in the Appendices \ref{Experiment 2} and \ref{Experiment 3}.

\subsection{Experiment 1: Provenance framework} \label{Experiment 4}

The experimental setup follows the methodology described in Section~\ref{System Description}. The pipeline consists of three main components: Relevancy Scorer, Context Item Selector, and Fact Checker. The Relevancy Scorer uses cross-encoder based models to rank context items based on their relevance to the given query. The Context Item Selector then selects top documents using either the TopK or TopP strategy. Finally, the Fact Checker evaluates the combined context to detect hallucinated content and returns a score. Results are presented in \autoref{table:exp1}.

\begin{table}[tb]
\centering
\caption{Results for Experiment 1: Provenance}
\label{table:exp1}
\begin{tblr}{
  width = \linewidth,
  colspec = {Q[340]Q[387]Q[204]},
  column{3} = {c},
  cell{3}{1} = {r=4}{},
  cell{7}{1} = {r=6}{},
  cell{13}{1} = {r=6}{},
  vlines,
  hline{1-3,7,13,19-21} = {-}{},
  hline{4-6,8-12,14-18} = {2-3}{},
}
\textbf{Dataset Type}                          & \textbf{Dataset}                    & \textbf{AUC} \\
\textbf{Paraphrase Detection}                  & \textbf{PAWS}                       & 0.94               \\
\textbf{Dialogue Generation}                   & \textbf{BEGIN}                      & 0.80               \\
                                               & \textbf{DIALFACT}                   & 0.92               \\
                                               & \textbf{Q2}                         & 0.86               \\
                                               & \textbf{HaluEval Dialogue}          & 0.69             \\
\textbf{Abstractive Summarization}             & \textbf{FRANK}                      & 0.89               \\
                                               & \textbf{MNBM}                       & 0.79               \\
                                               & \textbf{QAGS\_CNNDM}                & 0.76             \\
                                               & \textbf{QAGS\_XSUM}                 &0.80              \\
                                               & \textbf{Summ\_Eval}                 & 0.70              \\
                                               & \textbf{HaluEval Summarization}     & 0.66              \\
\textbf{Fact Verification}                     & \textbf{VITAMIN C}                  & 0.96              \\
                                               & \textbf{FEVER}                      & 0.92               \\
                                               & \textbf{Truthful\_QA}               & 0.59              \\
                                               & \textbf{MS\_MARCO}                  & 0.84               \\
                                               & \textbf{HaluBench}                  & 0.71               \\
                                               & \textbf{HaluEval QA}                & 0.74              \\
\textbf{Open Domain}                           & \textbf{HaluEval General}           & 0.54
\end{tblr}
\end{table}

\subsection{Experiment 2: Long context and multi-hop scenarios} \label{Experiment 5}

\begin{table}
\centering
\caption{Comparison of accuracies of different LLM-based methods in HaluBench \cite{ravi2024lynxopensourcehallucination} with \textit{Provenance}. The reported accuracy for \textit{Provenance} corresponds to Experiment 2, utilizing top\_k = 5 and the maximum aggregation logic.}
\label{table:halubench}
\begin{tblr}{
  width = \linewidth,
  colspec = {Q[421]Q[244]Q[242]},
  hlines,
  vlines,
}
\textbf{Models}      & \textbf{HaluBench} & \textbf{Model Size} \\
GPT-4o               & 87.9               & 1.7T                \\
GPT-4-Turbo          & 86.0               & 1.7T                \\
GPT-3.5-Turbo        & 62.2               & 175B                \\
Claude-3-Sonnet      & 84.5               & 70B                 \\
Claude-3-Haiku       & 68.9               & 20B                 \\
RAGAS Faithfulness   & 70.6               & 100B                \\
Mistral-Instruct-7B  & 78.3               & 7B                  \\
Llama-3-Instruct-8B  & 83.1               & 8B                  \\
Llama-3-Instruct-70B & 87.0               & 70B                 \\
LYNX (8B)            & 85.7               & 8B                  \\
LYNX (70B)           & \textbf{88.4}               & 70B                 \\
\textbf{\textit{Provenance}}  & 65.6     & 300M                
\end{tblr}
\label{table:4}
\end{table}

The experimental setup aligns with that of Section~\ref{Experiment 4}. In scenarios involving longer contexts and multi-hop scenarios, where answers span multiple context claims, as seen in HotpotQA \cite{yang2018hotpotqadatasetdiverseexplainable} and for some samples in HaluBench \cite{ravi2024lynxopensourcehallucination}, we aggregate the scores from the Fact Checker and weights from the Context Item Selector for each filtered \textit{source}. Results are presented in Table~\ref{table:hotpotqa}.

\begin{table}[tb]
\centering
\caption{Results from Experiment 2: Long context and multi-hop scenarios}
\label{table:hotpotqa}
\begin{tblr}{
  width = \linewidth,
  colspec = {Q[230]Q[233]Q[187]Q[187]},
  cell{3}{1} = {r=3}{}, 
  cell{6}{1} = {r=3}{}, 
  vlines,
  hline{1,3,6,9} = {-}{}, 
  hline{1-9} = {2-4}{},   
}
  {\\ \SetCell[c]{c} \textbf{Dataset}} & {\textbf{Selection}\\\textbf{Strategy}} & {\textbf{TopP 0.9}} & {\textbf{TopK 5}} \\
                        & {\textbf{Aggregation}} & {\textbf{AUC}}   & {\textbf{AUC}} \\
{\textbf{HotpotQA}}         & \textbf{min}                                       & 0.227                            & 0.440                          \\
                                        & \textbf{max}                                       & 0.809                            & 0.688                          \\
                                        & \textbf{weighted average}                              & 0.252                                & 0.372                          \\ 
\textbf{HaluBench }                     & \textbf{min}                                       & 0.645                            & 0.644                          \\
                                        & \textbf{max}                                       & 0.680                            & 0.714                          \\
                                        & \textbf{weighted average}                              & 0.664                                & 0.676                          
\end{tblr}
\end{table}

\begin{table}[tb]
\centering
\caption{Comparison of \textit{Provenance} accuracy to different models across various tasks presented in HaluEval \cite{li2023halueval}.}
\label{table:Provenance-accuracy-table}
\begin{tblr}{
  width = \linewidth,
  colspec = {Q[262]Q[112]Q[117]Q[174]Q[178]Q[133]},
  hlines,
  vlines,
}
\textbf{Models}      & \textbf{QA}    & {\textbf{Dia}\\\textbf{logue}} & {\textbf{Summa}\\\textbf{rization}} & \textbf{General~} & {\textbf{Model}\\\textbf{Size}} \\
ChatGPT              & 62.59          & \textbf{72.40}                 & 58.53                               & 79.44            & 175B                            \\
Claude 2             & \textbf{69.78} & 64.73                          & 57.75                               & 75.00            & 135B                            \\
Claude               & 67.60          & 64.83                          & 53.76                               & 73.88            & 130B                            \\
Davinci002           & 60.05          & 60.81                          & 47.77                               & \textbf{80.42}   & 6B                              \\
Davinci003           & 49.65          & 68.37                          & 48.07                               & 80.40            & 175B                            \\
GPT-3                & 49.21          & 50.02                          & 51.23                               & 72.72            & 13B                             \\
Llama 2              & 49.60          & 43.99                          & 49.55                               & 20.46            & 7B                              \\
ChatGLM              & 47.93          & 44.41                          & 48.57                               & 30.92            & 7B                              \\
Falcon               & 39.66          & 29.08                          & 42.71                               & 18.98            & 7B                              \\
Vicuna               & 60.34          & 46.35                          & 45.62                               & 19.48            & 7B                              \\
Alpaca               & 6.68           & 17.55                          & 20.63                               & 9.54             & 7B                              \\
\textbf{\textit{Provenance~}} & 67.48          & 62.97                          & \textbf{62.27*}                     & 56.70           & \textbf{300M}  
\end{tblr}
\end{table}

\section{Results}

\begin{figure}
    \centering
    \includegraphics[width=0.95\linewidth]{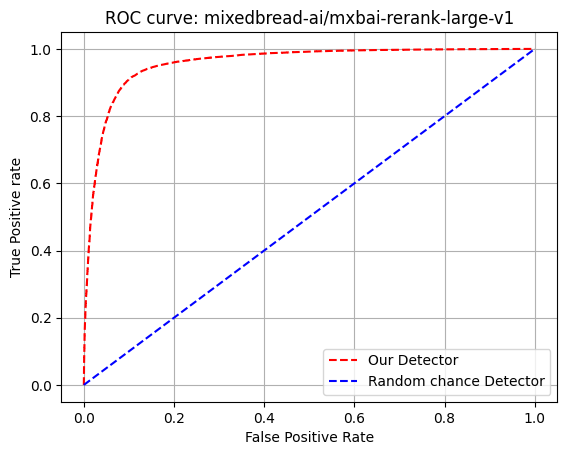}
    \caption{ROC curve for VITC task}
    \label{fig:roc-VITC}
\end{figure}
\begin{figure}
    \centering
    \includegraphics[width=0.95\linewidth]{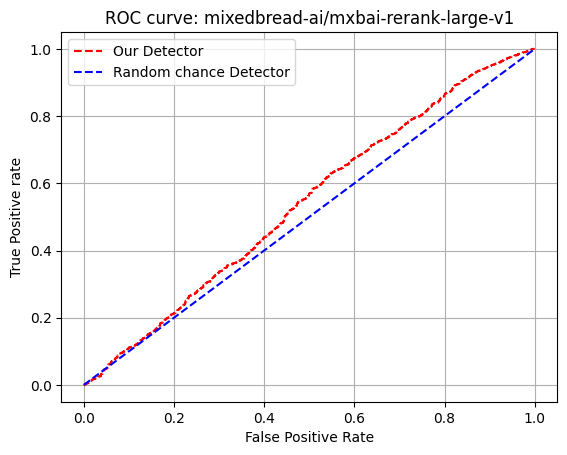}
    \caption{ROC curve for HALUEVAL-GENERAL task}
    \label{fig:roc-HALUEVAL-GENERAL}
\end{figure}

We report the ROC AUC of our system for all datasets mentioned in Section~\ref{data}. The ROC curves in Figures~\ref{fig:roc-VITC} and \ref{fig:roc-HALUEVAL-GENERAL} show the trade-off between false versus missed hallucination detections for the least and the most difficult of the test sets, respectively. Note that we did not reproduce the evaluations of the LLM-based methods listed in Tables~\ref{table:4} and~\ref{table:Provenance-accuracy-table}, and simply copied the results reported in the respective references.

\subsection{AUC Analysis}

Comparing our AUC scores with the TRUE dataset paper \cite{honovich2022true} in
Table~\ref{table:TRUE-accuracy-table}, our framework achieves the best AUC for \textbf{3 out of 7} datasets (DialFact, MNBM, and Q2).
Notably, the ANLI method \cite{honovich2022true}, which uses a 11B-parameter model, slightly outperforms ours on some datasets. Still, our model with
 \textbf{$\approx$ 300M} parameters shows competitive results with minimal differences: 0.4\% for FRANK and 3.4\% for QAGS\_XSUM, while performing \textbf{better by 2.9\%} for MNBM.

\subsection{Accuracy comparison}

Comparing accuracy scores from the HaluEval benchmark \cite{li2023halueval} in
Table~\ref{table:Provenance-accuracy-table}, \textit{Provenance} achieves the best accuracy on the summarization task, \textbf{surpassing ChatGPT by 3.74\%}, and is only 2.3\% behind Claude2 on the QA task, despite Claude 2 having 135B parameters.

Comparing accuracy scores from the HaluBench benchmark  \cite{ravi2024lynxopensourcehallucination} in
Table~\ref{table:halubench}, \textit{Provenance} is \textbf{surpassing GPT-3.5-Turbo by 3.38\%}, and is only 3.32\% behind Claude-3-Haiku, despite Claude-3-Haiku having two orders of magnitude more (20B) parameters.

\section{Conclusion}

We have presented Provenance, a practical approach to fact-checking of LLM output in RAG scenarios, based on light-weight cross-encoder models for relevance scoring and natural language inference. The factuality scoring takes the query into account when judging a generated answer against the retrieved information sources. 
Evaluation on a variety of open-source datasets shows our method to be effective for hallucination detection across a variety of tasks, at a model size that is a fraction of that of LLMs that are commonly used for this task.  We expect our method to make the fact-checking of LLM output more accessible and scalable, contributing to the reliability and trustworthiness of LLM-based applications.

\section*{Acknowledgments}

We thank Neha Gupta and Roberto Pieraccini for their support and advice in doing this research.

\newpage
\bibliography{custom}

\begin{table*}[t]
\centering
\caption{Baseline results from preliminary experiments on dot-product relevance scoring (Appendix~\ref{Experiment 1}), sources in temporal order (Appendix~\ref{Experiment 2}), and cosine similarity (Appendix~\ref{Experiment 3}).}
\label{table:combined}
\begin{tblr}{
  column{4} = {c},
  column{5} = {c},
  column{6} = {c},
  cell{3}{1} = {r=3}{},
  cell{6}{1} = {r=5}{},
  cell{11}{1} = {r=4}{},
  vlines,
  hline{1-3,6,11,15-16} = {-}{},
  hline{4-5,7-10,12-14} = {2-6}{},
}
\textbf{Data Type}              & \textbf{Dataset}     & \textbf{Sample Count} & \textbf{EXP-0.1 AUC} & \textbf{EXP-0.2 AUC} & \textbf{EXP-0.3 AUC} \\
{\textbf{Paraphrase} \\ \textbf{Detection}}   & \textbf{PAWS}        & 8000                  & 0.678                & 0.777                & 0.805                \\
{\textbf{Dialogue} \\ \textbf{Generation}}    & \textbf{BEGIN}       & 836                   & 0.632                & 0.749                & 0.749                \\
                                & \textbf{DialFact}    & 8689                  & 0.653                & 0.853                & 0.859                \\
                                & \textbf{Q2}          & 1088                  & 0.637                & 0.735                & 0.737                \\
{\textbf{Abstractive} \\ \textbf{Summarization}}  & \textbf{FRANK}    & 671                   & 0.452                & 0.720                & 0.790                \\
                                & \textbf{MNBM}        & 2500                  & 0.594                & 0.747                & 0.752                \\
                                & \textbf{QAGS\_CNNDM} & 235                   & 0.375                & 0.507                & 0.576                \\
                                & \textbf{QAGS\_XSUM}  & 239                   & 0.533                & 0.743                & 0.798                \\
                                & \textbf{Summ\_Eval}  & 1600                  & 0.447                & 0.546                & 0.639                \\
{\textbf{Fact} \\ \textbf{Verification}}     & \textbf{VITAMIN C}   & 63054                 & 0.607                & 0.813                & 0.825                \\
                                & \textbf{FEVER}       & 18209                 & 0.678                & 0.817                & 0.928                \\
                                & \textbf{TRUTHFUL\_QA}& 2976                  & 0.557                & 0.607                & 0.595                \\
                                & \textbf{MS\_MARCO}   & 504                   & 0.853                & 0.853                & 0.820                \\
\end{tblr}
\end{table*}

\appendix

\section{Appendix} \label{sec:appendix}
Here we report preliminary experiments to test the ability of a vector similarity approach in determining context relevance to a query. The principal conclusion of these experiments was that we needed better recall, switching to a cross-encoder scoring approach enabled this.

\subsection{Experiment 0.1: Dot-product scoring} \label{Experiment 1}

Our framework involved three main components: a sentence-tokenizer, a context filter, and a detector. The \textit{Spacy sentencizer}\footnote{https://spacy.io/api/sentencizer} tokenized the context paragraphs into sentences. These tokenized sentences, along with a formatted string combining the query and the answer ("The answer to the question \{query\} is \{answer\}."), are vectorized using a BERT-based model.\footnote{Available on huggingface as \href{https://huggingface.co/WhereIsAI/UAE-Large-V1}{WhereIsAI/UAE-Large-V1}}
A dot product is computed between each context sentence and the formatted string, selecting the most relevant context sentences based on the TopP selection strategy. These filtered context sentences and the formatted string are then passed to the NLI model%
\footnote{Available on huggingface as \href{https://huggingface.co/microsoft/deberta-v2-xxlarge-mnli}{microsoft/deberta-v2-xxlarge-mnli}}
to obtain the entailment scores. The ROC AUC score and ROC curve are derived from these entailment scores and ground-truth labels (0 for hallucination and 1 for correct answers). Results are presented in \autoref{table:combined}.

\subsection{Experiment 0.2: Temporal ordering of sources} \label{Experiment 2}

The experimental setup mirrors that of Appendix~\ref{Experiment 1}, with a minor modification in the context filter. Previously, the TopP selection strategy returned a list of relevant indices, which were directly mapped to context claims. In this updated approach, the filtered indices are sorted before mapping to ensure temporal order, so the context claim at index $n$ precedes the context claim at index $n+1$. The results are presented in \autoref{table:combined}.

\subsection{Experiment 0.3: Scoring with cosine similarity} \label{Experiment 3}
The experimental setup mirrors that of Appendix~\ref{Experiment 2}, but with a minor modification in the context filter. The vectorized context sentences and the formatted string are normalized to recreate cosine similarity for the dot product calculation. The results are presented in \autoref{table:combined}.

Columns~4 and~5 in Table~\ref{table:combined} show that maintaining the temporal order of filtered context claims enhances NLI model accuracy, especially for conversation-based use cases, yielding a \textbf{24.95\%} overall improvement in AUC scores.

Columns~5 and~6 in Table~\ref{table:combined} show that using cosine similarity results in a better threshold for the NLI model, with an overall \textbf{4.79\%} improvement in AUC scores.

Column~6 in Table~\ref{table:combined} and Column~3 in Table~\ref{table:4} demonstrate that the Relevancy Scorer with the Context Item Selector outperforms simple cosine similarity between context and query, leading to a \textbf{9.63\%} overall improvement in AUC scores.

\end{document}